\documentclass[twoside]{article}

\usepackage[accepted]{aistats2025}

\usepackage{hyperref}       % hyperlinks
\usepackage{url}            % simple URL typesetting
\usepackage{booktabs}       % professional-quality tables
\usepackage{amsfonts}       % blackboard math symbols
\usepackage{mathtools}
\usepackage{nicefrac}       % compact symbols for 1/2, etc.
\usepackage{microtype}      % microtypography
\usepackage{lipsum}		% Can be removed after putting your text content
\usepackage{graphicx}
\usepackage{xcolor}
\usepackage{doi}
\usepackage{amsthm}

\usepackage{algorithm}
\usepackage{algpseudocode}

\usepackage[round]{natbib}

\newtheorem{theorem}{Theorem}[section]

\DeclarePairedDelimiter\norm{\lVert}{\rVert}
\newcommand{\scalar}[2]{\langle #1, #2 \rangle}

\usepackage{subcaption}
\begin{document}



\twocolumn[

\aistatstitle{Function-Space MCMC for Bayesian Wide Neural Networks}

\aistatsauthor{ Lucia Pezzetti \And Stefano Favaro \And  Stefano Peluchetti }
%\thanks{Works done at Univerità di Torino}

\aistatsaddress{ ETH AI Center\\ lucia.pezzetti@ai.ethz.ch \And  University of Torino\\
and Collegio Carlo Alberto\\stefano.favaro@unito.it \And Sakana AI\\ stepelu@sakana.ai} 
]

\begin{abstract}
  Bayesian Neural Networks represent a fascinating confluence of deep learning and probabilistic reasoning, offering a compelling framework for understanding uncertainty in complex predictive models. In this paper, we investigate the use of the preconditioned Crank-Nicolson algorithm and its Langevin version to sample from a reparametrised posterior distribution of the neural network's weights, as the widths grow larger. In addition to being robust in the infinite-dimensional setting, we prove that the acceptance probabilities of the proposed algorithms approach 1 as the width of the network increases, independently of any stepsize tuning. Moreover, we examine and compare how the mixing speeds of the underdamped Langevin Monte Carlo, the preconditioned Crank-Nicolson and the preconditioned Crank-Nicolson Langevin samplers are influenced by changes in the network width in some real-world cases. Our findings suggest that, in wide Bayesian Neural Networks configurations, the preconditioned Crank-Nicolson algorithm allows for a scalable and more efficient sampling of the reparametrised posterior distribution, as also  evidenced by a higher effective sample size and improved diagnostic results compared with the other analysed algorithms.
\end{abstract}

\section{Introduction}

Neural Networks (NNs) have become very popular in the recent years, due to the significant results achieved in various artificial intelligence tasks from speech recognition and image classification, to stock market prediction, healthcare and weather forecasting. Despite this widespread applicability and indisputable success, NNs are hindered by intrinsic shortcomings. The large number of parameters that makes NNs powerful function approximations, is also the reason why NNs are prone to overfitting and overconfidence in their predictions. Moreover, the black-box nature of NNs makes interpretability hard. Bayesian Neural Networks (BNNs) address some of these limitations by providing a principled way to incorporate uncertainty into complex predictive models. At its heart lies the framework of Bayesian Inference, specifically, the weights or parameters of the network are considered to be random variables, instead of deterministic quantities, whose prior beliefs are updated as data are gathered according to the Bayesian learning rule \citep{khan2018fast, maddox2019simple, osawa2019practical, dusenberry2020efficient, daxberger2022bayesian, izmailov2021bayesian, Neal1996}. 

Despite all these promising advantages and some recent progress, mainly under Gaussian initializations of the weights, BNNs have reached far less popularity than NNs, due to higher computational requirements and limited theoretical understanding. One of the major theoretical challenges concerns the comprehension of the parameter-space behavior of BNNs, in spite of the function one. We therefore explore sampling from the posterior distriution of wide BNNs, focusing on understanding its behavior and properties from a parameter-space perspective. In function space, in fact, under Gaussian initializations of the network's weights, the distribution over the functions induced by wide BNNs has been proved to converge to the neural network Gaussian process (NNGP) limit \citep{matthews2018gaussian, lee2018deep, novak2020bayesian, garrigaalonso2019deep, yang2021tensor, hron2020exact, bracale2021largewidth}. Recently, \cite{hron_2022} provided a counterpart of this result in the parameter space. Specifically, they proved that, after a reparametrization $\phi = T^{-1}(\theta)$ of the flattened NN weights $\theta \in \mathbb{R}^d$, the posterior distribution $p(\phi | \mathcal{D})$ converges in the KL-divergence to the standard Gaussian distribution $\mathcal{N}(0, I_d)$ as the width of the BNN increases. This result not only gives a characterization of the wide-limit parameter distribution, but the closeness to the well-behaved standard normal distribution also suggests that an improvement in the mixing time of Markov chain Monte Carlo (MCMC) sampling procedures is possible. However, standard MCMC procedures are notably non-robust in the infinite-dimensional setting, since their proposal step needs to approach zero for the acceptance rate to remain constant, as the dimension of the space increases.

\subsection{Our contributions}

In this paper, we investigate and provide some theoretical guarantees for the use of MCMC algorithms that are stable under mesh refinement, namely the preconditioned Crank-Nicolson (pCN) and Crank-Nicolson Langevin (pCNL), to sample from the reparametrised posterior distribution of the weights as the widths of BNNs grow larger. Our primary objective is to advance the theoretical understanding of inference in BNNs, rather than to introduce a new state-of-the-art algorithm for practical applications. We show that the acceptance probability of both samplers, under reparametrization $T$, converges to 1 as the width of the BNN layers increases, independently of the chosen fixed stepsize. This result paves the way for a faster and more efficient sampling procedure in wide BNNs. Specifically, not only do the pCN and pCNL proposals avoid suffering from the curse of dimensionality, but the performance of the samplers intrinsically improves as the BNN becomes larger, without introducing any additional autocorrelation among the samplers. This lessens the need for longer burn-in periods for adequate convergence diagnostics, which affects standard MCMC methods.

The theoretical guarantees for the use of the pCN and pCNL proposals in the context of wide BNN configurations are empirically supported by a higher effective sample size and improved diagnostic results compared with the LMC algorithm. From a computational point of view, the pCN provides more scalability with respect to derivative-based MCMC algorithms and can scale well to large datasets and complex model. For this reason, the pCN method seems to effectively combine a small computational cost with a significant improvement in the quality of the collected samples, whereas the performance of the pCNL procedure is not striking enough to justify its high computational cost. Notably, the improvements in performance increase with the width size but are evident even far from the NNGP regime.

\subsection{Organization of the paper}

The paper is structured as follows. In Section \ref{sec1} and Section \ref{sec2} we present our methodologies based on the pCN and the pCNL proposals in wide BNNs. In Section \ref{sec3} we provide an empirical validation of our methodologies. In Section \ref{sec4} we discuss some directions for future research. The supplementary material contains the proofs of our results, and additional numerical experiments.

%%%%%%%%%%%%%%%%%%%%%%%%%%%%%%%%
%%%%%%%%%%%%%%%%%%%%%%%%%%%%%%%%
%%%%%%%%%%%%%%%%%%%%%%%%%%%%%%%%
%%%%%%%%%%%%%%%%%%%%%%%%%%%%%%%%

\section{Method}\label{sec1}

\subsection{Notations and preliminaries}
Let a dataset $\mathcal{D} = {(\mathbf{x}_1, \mathbf{y}_1), \dots, (\mathbf{x}_n, \mathbf{y}_n)}$, where $\mathbf{x}_i \in \mathbb{R}^m$ and $\mathbf{y}_i \in \mathbb{R}$. Let parameters $\theta$ belong to $\Theta \subset \mathbb{R}^D$, for some $D\geq1$. The prior distribution $p(\theta)$ represents our beliefs prior to observing the data, the beliefs are then updated accordingly to Bayes' rule as data are gathered:
\begin{displaymath}
p(\theta|\mathcal{D}) = \frac{p(\theta) p(\mathbf{y}| \theta, \mathbf{x})}{p(\mathbf{y}|\mathbf{x})},
\end{displaymath}
where $p(\mathbf{y} | \theta, \mathbf{x})$ is the likelihood function, i.e. the model for the data.

A BNN is a parametric function $f \coloneqq f_{\theta}$ that takes an input $\mathbf{x}$ and produces an output $\mathbf{y}$. To clarify the notation, we explicitly define the recursion for a fully connected L-hidden layer network with a linear readout layer, i.e.,
\begin{equation*}
f^{(0)}(\mathbf{x}) = \frac{\sigma^{(1)}_W}{\sqrt{d^{(0)}}}\mathbf{x}W^{(1)} + \sigma^{(1)}_b\mathbf{1}^{(0)}b^{(1)}
\end{equation*}
and for $l = 0, \dots, L$
\begin{equation*}
g^{(l)}(\mathbf{x}) = \psi(f^{(l)}(\mathbf{x})),
\end{equation*}
\begin{equation*}
f^{(l+1)}(\mathbf{x}) = \frac{\sigma^{(l+1)}_W}{\sqrt{d^{(l)}}}g^{(l)}(\mathbf{x})W^{(l+1)} + \sigma^{(l+1)}_b\mathbf{1}^lb^{(l+1)}
\end{equation*}

Thus, $\mathbf{y} = f^{L+1}(\mathbf{x})$ is the output of the network given the input $\mathbf{x}$. The (non-linear) activation function is denoted by $\psi$, $d^{(l)}$ denotes the width of the $l^{th}$ layer (with $d^{(0)}$ being the input size m) and $W^{(l)} \in \mathbb{R}^{d^{(l-1)}\times d^{(l)}}$ is its weight matrix. Without loss of generality, in the following we will not make explicit the presence of the bias $b^{(l)} \in \mathbb{R}^{d^{(l)}}$, though we include it in the weight matrix by adding a constant dimension to the outputs of the previous layer so to simplify the notation. The factor $\sigma^{l+1}_W/\sqrt{d^l}$ is to be attributed to the NTK parametrization \citep{jacot2020neural}, and it allows to consider an isotropic Gaussian prior for the weights, removing its usual dependence on the width. Precisely, we denote
\begin{displaymath}
\theta^{(l)} \in \mathbb{R}^{d^{(l)}}, \quad l=1, \dots, L+1, \quad\quad \theta^{(l)} \sim \mathcal{N}(0,I_{d^{(l)}})
\end{displaymath}
the flattened weights of the $l^{th}$ layer, bias included, assumed to be, a priori, independent standard Gaussian random variables. With $\theta \in \mathbb{R}^D$ we consider all the flattened weights $\theta = [\theta^{(l)}]_{l=0, \dots, L} \sim \mathcal{N}(0, I_D)$. We assume a Gaussian likelihood $p(\mathbf{y} | \theta, \mathbf{x}) \sim \mathcal{N}(f^{(L+1)}(\mathbf{x}), \sigma^2 I_D)$ for the observation variance $\sigma^2 > 0$. Moreover, we assume that $\left| \psi(x) \right|$ is bounded by a function of the form $\exp(C\| x \|^{2-\epsilon}+c)$, for some $C,c,\epsilon > 0$. This assumption still allows for functions that grow faster than exponential while remaining $L^1$ and $L^2$-integrable against Gaussian measures, with no constraint on their smoothness and encompasses almost all activations functions relevant in practice. See \cite{yang2021tensor} and references therein for similar assumptions on the activation function.

We are interested in theoretical guarantees for sampling from the posterior distribution of the weights as the width grows. \cite{hron_2022} proposed the following data-dependent reparametrization of the readout layer weights of BNNs:
\begin{equation}
\phi^{(l)} =
\begin{cases}
\Sigma^{-1/2}\left( \theta^{(l)} - \mu \right) & \quad l = L+1          \\[0.4cm]
\theta^{(l)}                                   & \quad \text{otherwise}
\end{cases}
\label{repar}
\end{equation}
where, by defining the quantity $\Psi = \frac{\sigma^{(L+1)}_W}{\sqrt{d^{(L)}}}g^{(L)}(\mathbf{x})$, we have
\begin{equation}
\Sigma = \left( I_{d^{L}} + \sigma^{-2} \Psi^T \Psi \right)^{-1}, \quad \mu = \sigma^{-2}\Sigma \Psi^Ty.
\label{cov_mean}
\end{equation}

The reparametrized posterior distribution, whose density function is denoted by $p(\phi | \mathcal{D})$, is shown to converge in the KL-divergence to a standard Gaussian $\mathcal{N}(0, I_D)$ as the width goes to infinity. In particular, the convergence to a simple isotropic Gaussian suggests potential improvements in the mixing speed of MCMC procedures, compared to sampling from the notably arduous BNN posterior. Nevertheless, standard MCMC algorithms, such as the Random Walk Metropolis-Hastings Algorithm (RW-MH) and the Metropolis Adjusted Langevin Algorithm (LMC), are notoriously ill-suited for the infinite-dimensional setting and must be carefully re-tuned as the dimension increases to avoid degeneracy in the acceptance probability. Indeed, the optimal scaling properties of the RW-MH and LMC algorithms have been derived in \cite{c56b4e38} and \cite{1c05a07d}, respectively. There, it is shown that the proposal movements need to vanish as the dimension of the parameter space grows unbounded for factorial target distributions.
Consequently, the efficiency of the RW-MH and LMC algorithms vanishes as well.

\subsection{Function-space MCMC}
\label{functionalMCMC}
Since our interest lies in analyzing the behavior of wide networks, we need methods that are robust in the infinite dimensional setting. A class of such MCMC algorithms have been early derived in \cite{Cotter_2013}. In this section, we focus on the pCN and pCNL methods and briefly contrast them with their traditional counterparts: the RW-MH and the LMC algorithms. The slight modifications introduced in pCN and pCNL made them robust and well-defined in high dimensional settings, contrasting the well known degeneracy of the standard MCMC acceptance probability as the dimension increases.

RH-MC employs a symmetric proposal, typically Gaussian, centered at the current state to explore the parameter space, i.e.,
\begin{equation*}
    v = u + \beta w, \quad w \sim \mathcal{N}(0, \mathcal{C}), \quad \beta \in (0,1]
\end{equation*}

Instead, the pCN algorithm relies on the following modification of the (standard) random walk proposal, i.e.,
\begin{equation}
v = \sqrt{1-\beta^2}u + \beta w, \quad w \sim \mathcal{N}(0, \mathcal{C}), \quad \beta \in (0,1]
\label{pcn_prop}
\end{equation}
where $\mathcal{C}$ is the covariance operator of the prior Gaussian measure. We refer to the work of \cite{Cotter_2013}, and references therein, for details on RH-MC and pCN proposals.

The pCN algorithm is reversible with respect to an arbitrary Gaussian measure, and hence admits such Gaussian measure as invariant distribution. Specifically, in our context, we set the invariant Gaussian measure to be equal to an isotropic Gaussian distribution (the prior). Accordingly, the acceptance probability reduces to:
\begin{equation}
a(v|u) = \min\{ 1, \exp(-\ell(u) + \ell(v))\}.
\label{pcn_acc}
\end{equation}

where $\ell$ is the log-likelihood. The pCN is in Algorithm \ref{alg: pcn}.

\begin{algorithm}
\caption{Preconditioned Crank-Nicolson (pCN) Algorithm}
\begin{algorithmic}[1]
\State Initialize $u^{(0)}$, set number of iterations $N$, choose $\beta \in (0, 1]$
\For{$n = 0$ to $N-1$}
    \State Propose $v = \sqrt{1-\beta^2} u^{(n)} + \beta \eta, \quad \eta \sim N(0, C)$
    \State Calculate acceptance probability $a(v| u^{(n)}) = \min\{ 1, \exp(-\ell(u^{(n)}) + \ell(v))\}$
    \State Draw $\eta \sim \text{Uniform}(0,1)$
    \If{$\eta \leq a(v| u^{(n)})$}
        \State Set $u^{(n+1)} = v$
    \Else
        \State Set $u^{(n+1)} = u^{(n)}$
    \EndIf
\EndFor
\end{algorithmic}
\label{alg: pcn}
\end{algorithm}

Despite being agnostic about which parts of the state space are more probable, the pCN proposal is, by construction, well-defined in the infinite-dimensional setting. This motivated us to study the procedure to sample from the above reparametrized weights posterior distribution. In our context, the proposal \ref{pcn_prop} is of the form
\begin{align*}
\phi^* &= \sqrt{1-\beta^2}\phi + \beta w \quad w \sim \mathcal{N}(0, I_D)\\ \implies &\quad q(\phi^* | \phi) = \mathcal{N}(\sqrt{1-\beta^2}\phi, \beta^2 I_D),    
\end{align*}

where $\beta \in [0, 1)$.
Now, by including this expression in the acceptance probability \ref{pcn_acc}, we can prove the following theorem.

\begin{theorem}
Consider the BNN model with the reparametrisation \ref{repar}. Then, the acceptance probability of the pCN algorithm to sample from the reparametrised weight posterior, for any $\beta \in [0, 1)$, converges to $1$ as the width of the network increases. If $d_{min}$ denotes the smallest among the network's layer widths, then as $d_{min} \to \infty$
\begin{equation*}
a(\phi^*|\phi)  =  \min\left\{ 1, \frac{p(\phi^* | \mathcal{D})q(\phi | \phi^*)}{p(\phi | \mathcal{D})q(\phi^* | \phi)} \right\} \to 1
\end{equation*}
\label{Thm1}
\end{theorem}

We provide a high-level sketch of the proof of Theorem \ref{Thm1}. We refer to Appendix \ref{proof_acc} for the full proof of Theorem \ref{Thm1}.

\begin{proof}
The acceptance probability for the Metropolis-Hastings step is given by:
\begin{equation*}
a(\phi^* | \phi) = \min \left\{ 1, \frac{p(\phi^* | \mathcal{D}) q(\phi | \phi^*)}{p(\phi | \mathcal{D}) q(\phi^* | \phi)} \right\}.
\end{equation*}

In particular, by making use of the reparametrization \(\phi = T^{-1}(\theta)\), the posterior distribution \(p(\phi | \mathcal{D})\) converges in KL-divergence to a standard Gaussian \(\mathcal{N}(0, I_d)\) in the infinite-width limit. Furthermore, the pCN proposal is designed to preserve Gaussian measures, meaning that:
\begin{equation*}
    q(\phi^* | \phi) = \mathcal{N} \left( \sqrt{1 - \beta^2} \phi, \beta^2 I_d \right).
\end{equation*}
Using this, the ratio of proposal densities simplifies to the following:
\begin{equation*}
    \frac{q(\phi | \phi^*)}{q(\phi^* | \phi)} = \exp \left( \frac{1}{2} ( \| \phi^* \|^2 - \| \phi \|^2 ) \right).
\end{equation*}

Along similar lines, we can express the posterior probability ratio in terms of the empirical NNGP kernel \(\hat{K}_{\sigma^2}\), which converges to a constant in the infinite-width limit:
\begin{equation}
    \frac{p(\phi^* | \mathcal{D})}{p(\phi | \mathcal{D})} \propto \frac{\sqrt{\det(\hat{K}^*_{\sigma^2})} \exp \left( \frac{1}{2} y^\top (\hat{K}^*_{\sigma^2})^{-1} y \right)}{\sqrt{\det(\hat{K}_{\sigma^2})} \exp \left( \frac{1}{2} y^\top (\hat{K}_{\sigma^2})^{-1} y \right)}.
\end{equation}

Since \(\hat{K}_{\sigma^2} \to K_{\sigma^2}\) in the limit, the ratio converges to 1, implying that:
\begin{equation*}
    a(\phi^* | \phi) \to 1 \quad \text{as} \quad d_{\min} \to \infty.
\end{equation*}
\end{proof}

We address here the choice of the white noise reference measure. A common assumption in infinite-dimensional sampling theory is that the reference Gaussian measure should have a trace-class covariance operator, as recently developed in the work of \cite{sell2022traceclassgaussianpriorsbayesian}, meaning that the sum of its eigenvalues is finite. 
However, in our work, we select a reference measure with an identity covariance operator \( C = I \), leading to a white noise prior.
From an intuitive standpoint, our approach does not rely on an explicit basis representation as in \cite{Cotter_2013}. Instead, we consider an alternative formulation where the posterior distribution itself converges to the base distribution in the limit of increasing width. To clarify this, let us consider a simple example.

Let \( \mathcal{N}(n) \) denote a centered Gaussian distribution with identity covariance matrix in \( n \) dimensions. When the proposal is constructed via pCN, it takes the following form:

\begin{equation}
    \theta' \gets \sqrt{1 - \beta^2} \theta + \beta \epsilon, \quad \epsilon \sim \mathcal{N}(n),
\end{equation}

for \( \beta \in (0,1] \). If \( \theta \sim \mathcal{N}(n) \), then we also have \( \theta' \sim \mathcal{N}(n) \), ensuring that the proposal preserves the reference measure. 

In contrast, when using the Metropolis Random Walk proposal:

\begin{equation}
    \theta' \gets \theta + \beta \epsilon, \quad \epsilon \sim \mathcal{N}(n),
\end{equation}

we observe that even if \( \theta \sim \mathcal{N}(n) \), the new sample \( \theta' \) is not distributed as \( \mathcal{N}(n) \). Accordingly, such a critical difference leads to distinct acceptance rate behaviors as \( n \to \infty \).

The LMC algorithm incorporates gradient information from the log-posterior distribution in the proposal mechanism to guide the sampling process. More precisely, this method involves simulating Langevin diffusion such that the solution to the time evolution equation is a stationary distribution that equals the target density (in Bayesian contexts, the posterior distribution). This approach is particularly effective in exploring complex, high-dimensional probability distributions by making proposals that are more informed and therefore likely to be accepted. The mathematical expression of the proposal is:
\begin{equation*}
    v = u + \frac{\epsilon^2}{2} \nabla \log p(u|D) + \epsilon w, \quad w \sim N(0, I)
\end{equation*}
where $v$ is the current position of the algorithm in the parameter space, and $\epsilon$ is the stepsize,  which is a tuning parameter that allows to control the scale of the updates. In particular, $\nabla \log p(\phi|D)$ denotes the gradient of the log-posterior density function, which directly provides the direction towards higher probability densities.

While LMC provides an efficient way to explore the parameter space, it can still struggle in scenarios where the parameter space is high-dimensional. This led to the development of the pCNL algorithm, which introduces little modifications to better handle these challenges. More precisely, its proposal is derived by discretization of a stochastic partial differential equation (SPDE) which is invariant for target measure and is given by:
    \begin{equation}
    v = \frac{1}{2+\delta} \left[(2-\delta)u + 2\delta\mathcal{C}\mathcal{D}\ell(u) + \sqrt{8\delta} w \right],
    \label{pcnl_prop}
    \end{equation}
where $w \sim \mathcal{N}(0,\mathcal{C})$, $\delta \in (0,2)$, $\mathcal{C}$ is the covariance operator of the Gaussian prior measure, $\ell$ is the log-likelihood, and $\mathcal{D}\ell$ denotes its Fréchet derivative. We refer to Appendix \ref{frechet} for more details on Fréchet derivative. Note that the parameter $\delta$ is directly related to the pCN stepsize $\beta$ by the relationship $\beta^2 = 8\delta / (2+\delta)^2$. Throughout the paper, we refer to the stepsize as $\beta$, and we keep into account this relationship when dealing with the pCNL proposal. When the pCNL is used as a proposal for a MH scheme, the acceptance probability is given by 
\begin{equation}
    a(v|u) = min\{1, \exp(\rho(u,v)) - \rho(v,u)\},
\label{pcnl_acc}
\end{equation} where

\begin{align*}
    \rho(u,v) &= - \ell(u) -\frac{1}{2} \scalar{v-u}{\mathcal{D}\ell(u)}\\ &- \frac{\delta}{4} \scalar{u+v}{\mathcal{D}\ell(u)} + \frac{\delta}{4}\| \sqrt{\mathcal{C}}\mathcal{D}\ell(u) \|^2
\end{align*}

In particular, these modifications ensure that pCNL retains the efficiency of LMC in terms of informed proposals while enhancing its applicability and robustness in more challenging high-dimensional settings. The steps of the pCNL procedures are presented in Algorithm \ref{alg: pcnl}.

\begin{algorithm}
\caption{Preconditioned Crank-Nicolson Langevin (pCNL) Algorithm}
\begin{algorithmic}[1]
\State Initialize $u^{(0)}$, set number of iterations $N$, choose $\delta \in (0,2)$
\For{$n = 0$ to $N-1$}
    % \State Propose $v = \frac{1}{2+\delta} \left[(2-\delta)u^{(n)} + 2\delta\mathcal{C}\mathcal{D}\ell(u^{(n)}) + \sqrt{8\delta} \mathcal{N}(0,\mathcal{C})\right]$
    \State Draw $v = \frac{(2-\delta)}{2+\delta}u^{(n)} + \frac{2\delta\mathcal{C}}{2+\delta}\mathcal{D}\ell(u^{(n)}) + \frac{\sqrt{8\delta}}{2+\delta} \eta, \newline  \eta \sim \mathcal{N}(0,\mathcal{C})$
    \State Calculate acceptance probability $a(v| u^{(n)}) = \min\left\{1, \frac{p(v|D)}{p(u^{(n)}|D)}\right\}$
    \State Draw $\eta \sim \text{Uniform}(0,1)$
    \If{$\eta \leq a(v| u^{(n)})$}
        \State Set $u^{(n+1)} = v$
    \Else
        \State Set $u^{(n+1)} = u^{(n)}$
    \EndIf
\EndFor
\end{algorithmic}
\label{alg: pcnl}
\end{algorithm}

It is therefore natural to ask whether also the acceptance probability of the pCNL algorithm converges to one as the width of the neural network increases to infinity.  

Firstly recall the pCNL proposal \ref{pcnl_prop} that, using the notation introduced with the reparametrisation \ref{repar}, can be written as:
\begin{equation*}
    \phi^* = \frac{1}{2+\delta} \left[(2-\delta)\phi + 2\delta\mathcal{C}\mathcal{D}\ell(\phi) + \sqrt{8\delta} Z \right],
\end{equation*}
where $Z \sim \mathcal{N}(0,\mathcal{C})$ and $\delta \in (0,2)$.

Then, when the pCNL proposal is applied in the MH procedure, the acceptance probability \ref{pcnl_acc} has the following expression:
$$acc_{pCNL} = min\left\{ 1, exp(\rho(\phi,\phi^*) - \rho(\phi^*,\phi)) \right\}$$

where
\begin{align*}
    \rho(\phi,\phi^*) &= - \ell(\phi) - \frac{1}{2}\langle\phi^*-\phi, \mathcal{D}\ell(\phi)\rangle\\ &- \frac{\delta}{4} \langle\phi+\phi^*, \mathcal{D}\ell(\phi)\rangle + \frac{\delta}{4} \norm{\sqrt{C}\mathcal{D}\ell(\phi)}^2
\end{align*}

and, in our case, $C = I$.

%The next theorem provides the convergence of the acceptance probability of the pCNL algorithm as the network width increases.
\begin{theorem}
    Consider the BNN model with the reparametrisation \ref{repar}. Then, the acceptance probability of the pCNL algorithm to sample from the reparametrised weight posterior, 
    for any $\delta \in (0,2)$, converges to $1$ as the width of the network increases. That is to say, if $d_{min}$ denotes the smallest among the network's layer widths, then as $d_{min} \to \infty$
    \begin{equation*}
    a(\phi^*|\phi)  = \min\left\{ 1, \frac{p(\phi^* | \mathcal{D})q(\phi | \phi^*)}{p(\phi | \mathcal{D})q(\phi^* | \phi)} \right\} \to 1
    \end{equation*}
    \label{Thm2}
\end{theorem}

The proof of Theorem \ref{Thm2} is along lines similar to the proof of Theorem \ref{Thm1}. See Appendix \ref{proof_thm2} for the proof of Theorem \ref{Thm2}.

\section{Marginal-conditional decomposition}\label{sec2}
As an alternative approach to what presented in Theorem \ref{Thm1}, we consider marginalizing the weights of the network's final layer and perform the sampling procedure only on the weights of the network's inner layers. This idea is effective since it acknowledges that exact sampling can be performed from the posterior distribution of the reparametrized weights of the last layer, once the weights from all preceding layers are known, i.e.,
\begin{align*}
p(\phi | \mathcal{D}) & = p(\phi^{(L+1)} | \phi^{(\le L)}, \mathcal{D}) p(\phi^{(\le L)} | \mathcal{D})     \\
& = p(\phi^{(L+1)} | \theta^{(\le L)}, \mathcal{D}) p(\theta^{(\le L)} | \mathcal{D}),
\end{align*}
where the last equality follows directly from the reparametrisation definition and from the fact that  $p(\phi^{(L+1)} | \theta^{(\le L)}, \mathcal{D}) \sim \mathcal{N}(0, I_{d^{(L)}})$ for any fixed value of $\theta^{(\le L)}$. The idea we develop is then to simply perform pCN sampling on the posterior distribution over the inner-layers weights $\pi(\theta^{\le L} | \mathcal{D})$. Therefore, once the samples
\begin{displaymath}
\left[\theta^{(\le L)}_i\right]_{i = 1, \dots, n}
\end{displaymath}
have been collected, we draw, $\forall i$, $\phi^{(L+1)}_i \sim \mathcal{N}(0, I_{d^{(L)}})$, to obtain a sample of the full posterior distribution of the reparametrised weights. On the pCN algorithm, the next theorem is the natural counterpart of Theorem \ref{Thm1}.

\begin{theorem}
Consider the  BNN model with reparametrisation \ref{repar}, and set $\theta^{(\le L)} = W$. By the chain rule, $p(\phi | \mathcal{D}) = p(\phi^{(L+1)} | W, \mathcal{D}) p(W | \mathcal{D})$. Then, the acceptance probability of the pCN algorithm, for any $\beta \in [0, 1)$, applied to $p(W | \mathcal{D})$ converges to $1$ as the width of the network increases. If $d_{min}$ denotes the smallest among the network's layer widths, then as $d_{min} \to \infty$
\begin{equation*}
a(\phi^*|\phi)  = \min\left\{ 1, \frac{p(W^* | \mathcal{D})q(W | W^*)}{p(\phi | \mathcal{D})q(W^* | W)} \right\} \to 1
\end{equation*}
\label{Thm3}
\end{theorem}

See Appendix \ref{proof_acc_margin} for the proof of Theorem \ref{Thm3}.

\section{Numerical experiments}\label{sec3}

We test our framework to validate the theory and highlight its applicability. In the setting of Theorems \ref{Thm1} and \ref{Thm2}, we compare the performance of the pCN algorithm, the underdamped LMC sampler and the pCNL method on the reparametrized posterior of a fully connected feed-forward BNN. We replicate the setting used by \cite{hron_2022}: the {\fontfamily{qcr}\selectfont CIFAR-10} dataset, with one-hot encoding and $-1/10$ label shifting, is used in all the experiments, the likelihood is Gaussian with a fixed standard deviation of $\sigma = 0.1$ and sample thinning is consistently applied. A fair comparison between the different samplers is achieved by equalizing their stepsize $\beta$, that in our context always refers to the coefficient multiplying the noise in the proposal step. The experiments were conducted using a machine equipped with an NVIDIA A40 GPU, 64 GB of RAM, and 8 vCPUs. The code for experiments is available at \href{https://github.com/lucia-pezzetti/Function-Space-MCMC-for-Wide-BNNs}{github.com/lucia-pezzetti/Function-Space-MCMC-for-Wide-BNNs}.

\subsection{Acceptance rate of convergence}
\begin{figure*}
\centering
\includegraphics[width=\textwidth]{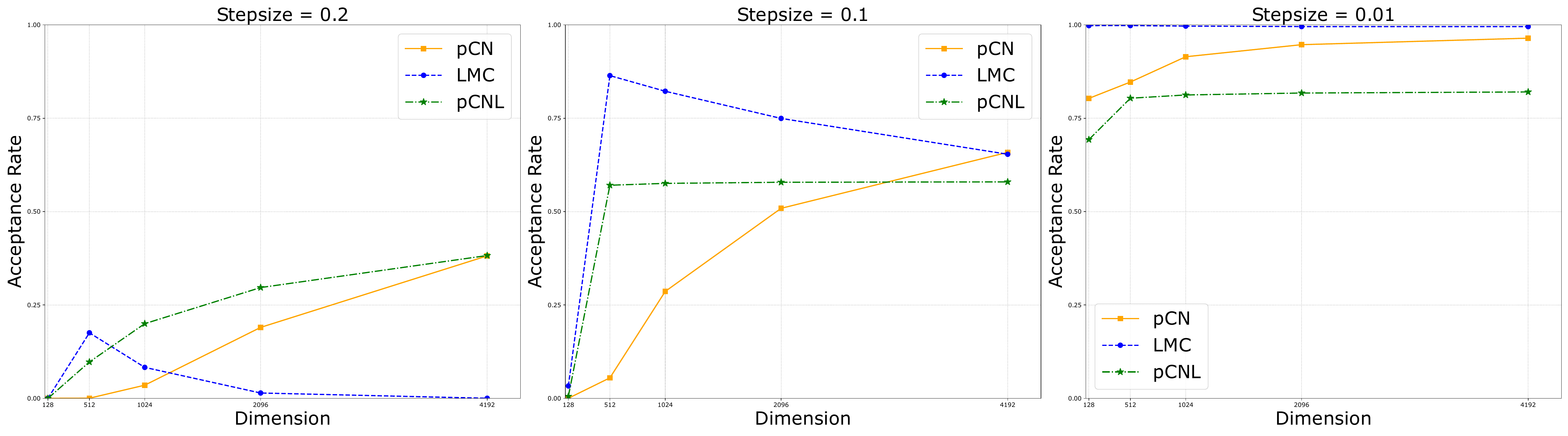}
\caption{Comparison at different stepsizes ($\beta = 0.2, 0.1, 0.01$) of the acceptance probability obtained using: i. the underdamped LMC algorithm (or Metropolis Adjusted Langevin Algorithm: LMC); ii. the pCN algorithm; iii. the pCNL method. The neural network architecture used is a fully-connected with one hidden layer, and layer width that varies among the following values: ${512, 1024, 2048, 4096, 8192}$. The {\fontfamily{qcr}\selectfont CIFAR-10} dataset is used, with the sample size fixed at $n=256$. The acceptance rate of the pCN increases steadily as the width of the BNN grows with the stepsize $\beta$, suggesting improved performance in wide BNNs and empirically confirming our theoretical analysis. The pCNL algorithm shows a similar trend in its acceptance rate, outperforming the other samples. In contrast, the LMC initially shows generally a deterioration in its acceptance rate as the width of the BNN increases, reflecting the sampler's non-robustness in high-dimensional settings.}
\label{fig:acc_rate_comp}
\end{figure*}

The first result we report in Figure \ref{fig:acc_rate_comp} is the numerical illustration of the behavior of the pCN, LMC and pCNL acceptance rates as the width of the BNN grows large. To be more precise, the acceptance probabilities of the underdamped LMC the pCN and the pCNL, applied to the posterior of the reparametrized weights, are plotted for a FCN with 1 hidden layer. For each subplot a fixed stepsize, respectively $\beta = 0.2$, $\beta = 0.1$ and $\beta = 0.01$, has been used in the proposal step. The figure clearly shows that, for each $\beta$, the acceptance rate of the pCN algorithm steadily increases with the layer width, empirically showcasing the theoretical result presented in Theorem \ref{Thm1}.

As expected from Theorem \ref{Thm2}, a similar trend is also observed for the pCNL algorithm. In particular, an interesting aspect that emerges is that the introduction of the Langevin dynamics in the pCN algorithm seems to gradually become less decisive as the width of the network increases, as showcased by the reduction, or even inversion, in the gap between the pCN and the pCNL acceptance rates. A plausible hypothesis for this behaviour is connected to the loss of representative learning in NNGPs, as the NNGP kernel is a deterministic function of the inputs and this can make the pCN preferable in large wisth scenarios, with respect to the introduction of some forms of likelihood information \citep{MacKay1998IntroductionTG, aitchison2020biggerbetterfiniteinfinite, yang2023theoryrepresentationlearninggives}. 
Finally, the profile of the LMC acceptance rates confirms and reflects the reason why we started our investigations: the MCMC methods' non-robustness in high-dimensional settings leads to a progressive decline of its acceptance rate as the width increases, which can only be compensated for by decreasing the method stepsize $\beta$.

\label{subsec_acc_rate_conv}

We then proceed by analyzing the impact that the guarantees obtained on the acceptance rate of the pCN and pCNL algorithms have in terms of improvements in the quality of the samplers using two key diagnostic tools: the effective sample size and the Gelman-Rubin statistic.

\subsection{Diagnostic: effective sample size}
We assess the performance of the MCMC using the effective sample size (ESS). For an MCMC-based estimator, the ESS estimates the number of independent samples that are equivalent, in terms of their standard variance, to the correlated Markov chain samples collected. We use TensorFlow Probability’s built-in function, \texttt{tfp.mcmc.effective\_sample\_size}, to compute
\begin{equation*}
ESS(N) =  \frac{N}{1 + 2 \sum_{i=1}^{N-1} (1-\frac{i}{N})R_{i}},
\end{equation*}
where the resulting sequence of auto-correlations is truncated after the first appearance of a term that is less than zero \citep{dillon2017tensorflow}. The per-step ESS will be calculated by dividing the above expression by $N$.

The ESS is a crucial metric to evaluate the performance of a MCMC procedure, as it helps to assess the quality of the samples and how well the Markov chain explores the target distribution by giving a measure of the autocorrelation among the collected samples. However, it does not give any information about the convergence of the Markov chain to the target distribution. To address this and have a more complete picture of the efficacy of the samplers, in Appendix \ref{gelman_rubin} we report the results obtained for the Gelman-Rubin statistic, $\hat{R}$. Another delicate point is that the ESS only provides useful insights if the burn-in period is adequately chosen. We empirically justify our choice using the trace plots of the principal components of the iterates in Appendix \ref{trace_plots}.

\subsection{Real world application}
\begin{figure*}[ht]
    \centering
    \begin{subfigure}{0.32\textwidth}
    \centering
    \includegraphics[width=\textwidth]{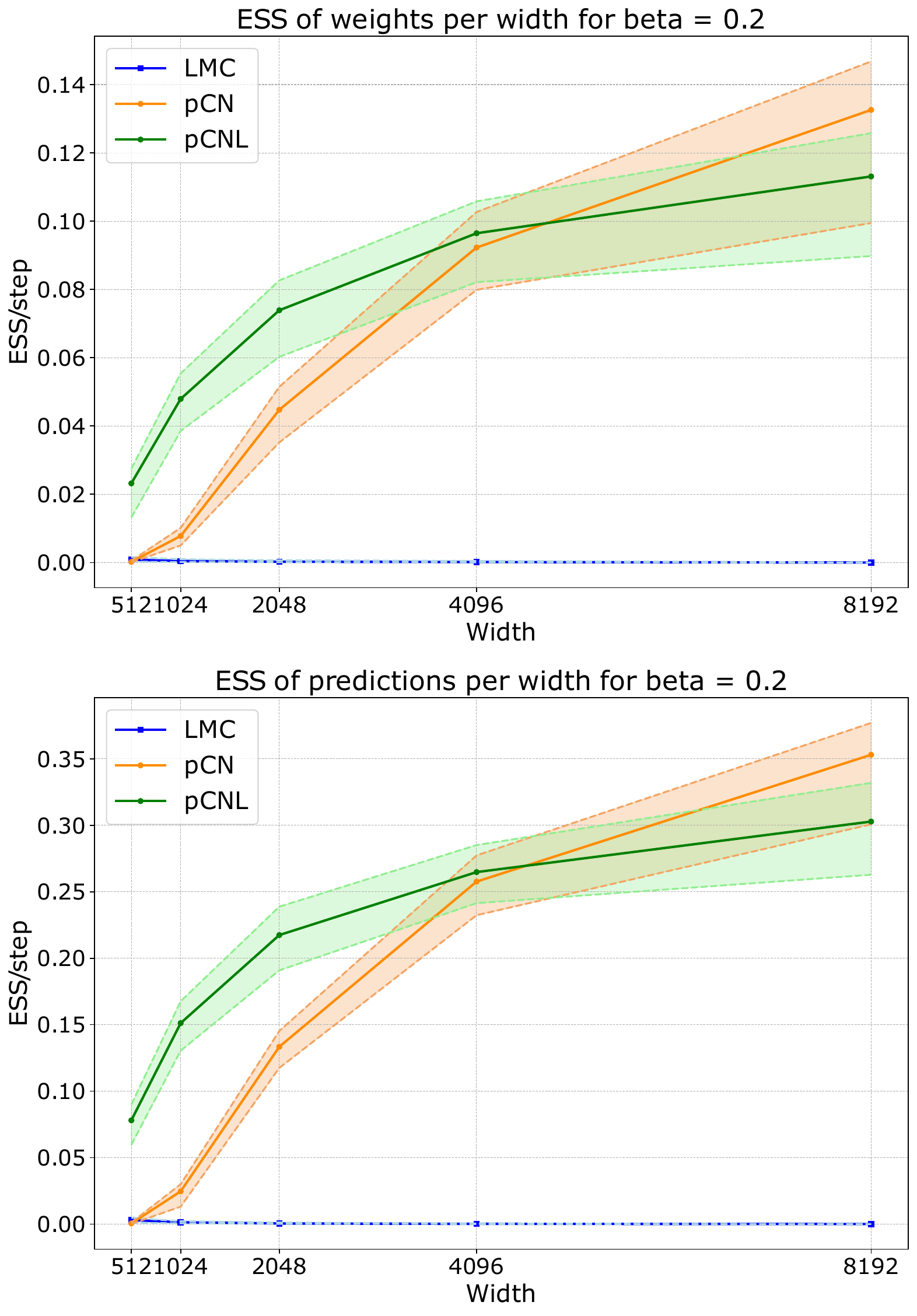}
    \end{subfigure}%
    %add desired spacing between images, e. g. ~, \quad, \qquad etc.
    \,
    \begin{subfigure}{0.32\textwidth}
    \centering
    \includegraphics[width=\textwidth]{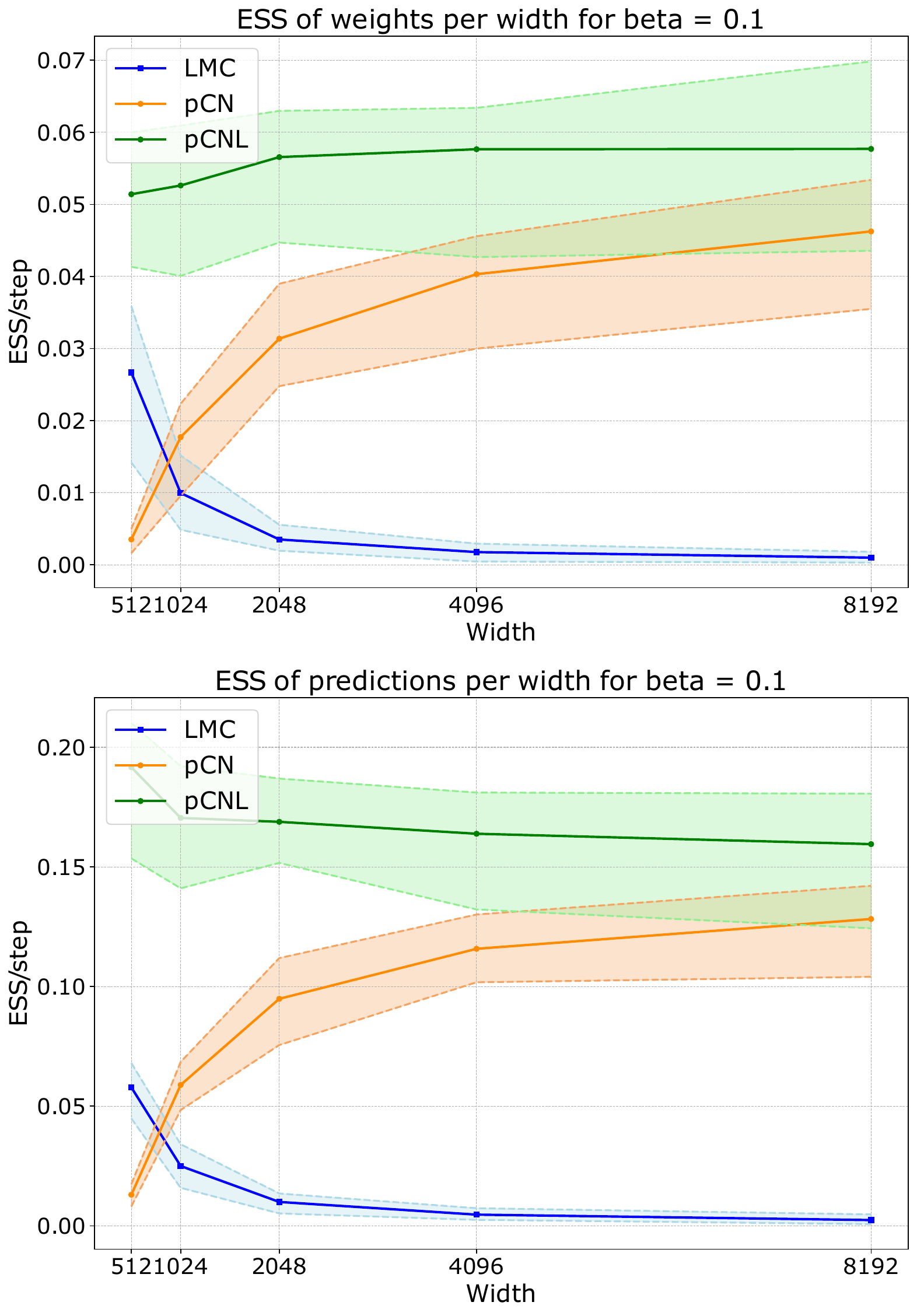}
    \end{subfigure}
    \,
    \begin{subfigure}{0.329\textwidth}
        \centering
        \includegraphics[width=\textwidth]{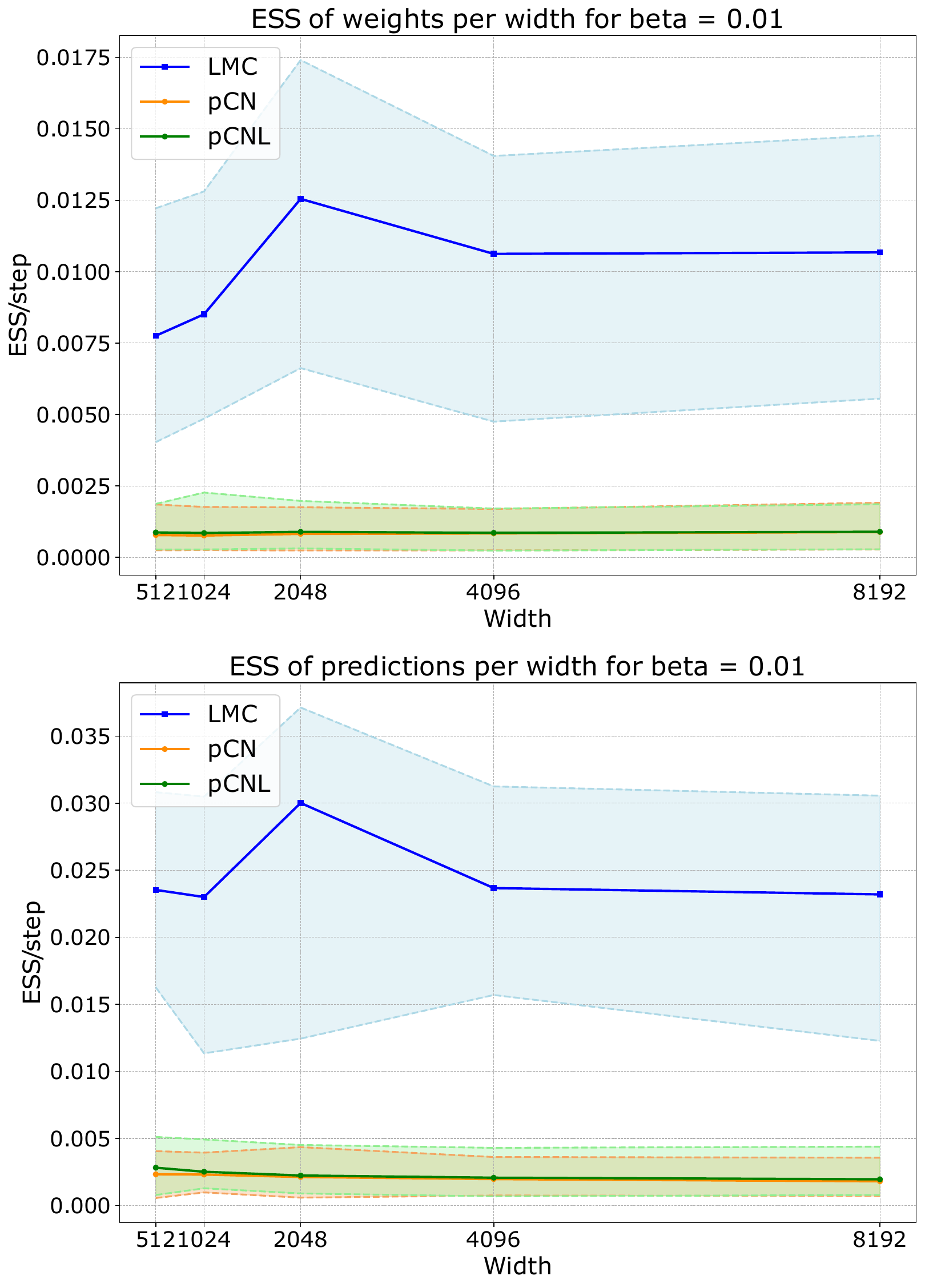}
        \end{subfigure}

    \caption{ESS analysis of the LMC, pCN and pCNL algorithms as a function of the 1-layer FCN's width for stepsizes $\beta = 0.2$ (left), $\beta = 0.1$ (middle) and $\beta = 0.01$ (right). 
    The solid lines represent the average per-step ESS, whereas the shaded areas indicate the variability of the per-step ESS delineated by its minimum and maximum 
    values. The setting used in the experiments is the same as the setting of Figure \ref{fig:acc_rate_comp}: the layer width of the BNN varies among the following values: 
    $\{512,\, 1024,\, 2048,\, 4096,\, 8192\}$. The {\fontfamily{qcr}\selectfont CIFAR-10} dataset is used, with sample size fixed at $n=256$.
    The poor LMC performance reflects the fact that standard MCMC procedures are ill-posed in high-dimensional settings. In contrast, the pCN and pCNL samplers demonstrate constant growth in ESS as the network width increases, indicating that enhancements in acceptance rate contribute positively to efficiency and performance. Finally, the smallest stepsize, 
    $\beta = 0.01$, heavily affects the behavior of both algorithms, introducing high autocorrelation among the samples and affecting their quality.}
    \label{fig:ess}
\end{figure*}

We examine and compare how the efficiency of the LMC, the pCN and the pCNL samplers are influenced by changes in network width, using per-step ESS as the performance metric. Our analysis was conducted using a single chain over one million steps, excluding the first 20,000 steps as burn-in for each configuration tested, and we thin by collecting every 25th step from the chain. To isolate the effects of network width increase, in all experiments, we used a single-layer, fully connected feed-forward neural network with GELU activation functions \citep{hendrycks2023gaussian}. We adhered to the weight and bias scaling described in Section 1.1 of \cite{hron_2022}, setting the scaling factors to $\sigma_{W}^2 = 2$ and $\sigma_b^2 = 0.01$ across all layers except for the output layer, where $\sigma_{W}^2$ was set to 1 to maintain unit variance of the network's initial predictions under the prior distribution.

In this setting, we focused on evaluating the samplers' performance as the network width increased, seeking empirical validation of our theory. We fixed the dataset size at n=256 and allowed the network width to range from 512 to 8192, increasing accordingly to powers of 2. Note that the reparametrization \ref{repar} was used for all samplers.

Empirical findings are shown in Figure \ref{fig:ess}, which compares the efficiency of the samplers using the per-step ESS metric across various widths. Specifically, the solid lines in Figure \ref{fig:ess} represent the average ESS per step for each width, while the shaded regions indicate the minimum and maximum values over the whole run. The reported plots correspond to the stepsizes $\beta \in \{0.2, 0.1, 0.01\}$.

As expected, reducing the MCMC algorithms' stepsizes in order to maintain high acceptance rates leads to a strong autocorrelation among the samples and, consequently, an extremely low per-step ESS. This emphases the unfeasibility of using standard MCMC approaches in large width settings and hence also provides an empirical cofirmation of the motivations behind this study.

More meaningful results, in terms of ESS, are achieved for the $\beta=0.1$ and $\beta=0.2$ stepsizes. Specifically, we can see that the acceptance rates' trends visible in Figure \ref{fig:acc_rate_comp} are quite faithfully preserved in the per-step ESS plots. The ESS of both the pCN and pCNL procedures increases over time and, since all the parameters except the network width are kept fixed, we can argue that this can be considered as a direct implication of our theoretical guarantees on the methods acceptance rates. More precisely, Theorems \ref{Thm1} and \ref{Thm2} says that as the width of the network increases, the acceptance probabilities associated to the pCN and pCNL proposals converge towards 1 "naturally", i.e. without any needing of parameter tuning. In other words these results capture an intrinsic property of the networks and do not affect the autocorrelation among the samples. As a consequence, it is reasonable and intuitive that a higher acceptance rate leads to a higher ESS. Once again, as for the analysis of the acceptance rates, the performances of the pCN algorithm seems to eventually reach the ones of the pCNL in wide BNNs. On the contrary, the LMC algorithm presents very low ESS for both stepsizes $\beta$. If this behavior is not surprising when considerig the larger $\beta$, due to the sampler suboptimal acceptance rate, the discussion is more intresting for the stepsize $\beta=0.1$. In this case, in fact, it is evident the quick deterioration in performance associated to the well-known MCMC methods' non-robustness in high-dimensional settings. Finally, we stress that discussing the ergodicity of  These empirical findings are further supported and validated in Appendix \ref{additional_experiments} by providing an analysis of the Gelman-Rubin statistic and the trace plots of the principal components of the iterates.

In general, the results of the experiments are consistent with the theoretical analysis. We argue that the pCN sampler emerges as the most appropriate method for sampling from the reparametrized posterior of wide BNNs, since it combines good performances with a reduced complexity. The pCNL method, in fact, does not have enough meaningful outcomes to justify its heavy computational demands. Furthermore, the experiments underscore the weaknesses of standard MCMC algorithms in this setting and the need to look at different approaches.

Finally, it is worth noting that during this work we focused only on very large width regimes for the network. In lower dimensional settings, instead, may be worth tuning the stepsize of classic MCMC methods or employing the pCNL algorithm in order to guarantee the best results.

\section{Discussion}\label{sec4}

We investigated the pCN and pCNL algorithms in sampling wide BNNs, providing insights into their scalability and efficiency, in comparison to the underdamped LMC sampler. Our theoretical findings, grounded in the understanding of the reparameterized posterior distribution, demonstrated that the acceptance probability of the pCN and pCNL proposals converges to $1$ as the network width goes to infinity, thus ensuring efficient sampling. Empirical results further corroborated this theoretical prediction, showing that, when applied to fully connected networks with large widths, these function-space MCMC algorithms consistently achieved higher per-step ESS than LMC, especially as dimensionality increased. This finding is significant as it underscores the advantage of these approaches in wide BNNs, where standard MCMCs often struggle due to non-robustness and high autocorrelation among samples. 

More precisely, our results indicate that the pCN sampler, within the proposed framework, offers a reliable and scalable solution for Bayesian inference in the context of wide neural networks, ultimately leading to better diagnostic performance, as well as an improved computational efficiency. However, for neural networks with smaller widths, the pCN method appears too naive, and it is significantly outperformed by its Langevin counterpart. 

Both the pCN and the pCNL show promise as complementary alternatives to standard MCMC algorithms for wide BNNs. While the pCN algorithm is scalable and highly effective in very large networks, its performance decreases sharply as the number of neurons decreases. The gradient information within the MH proposal shows potential as a solution to this limitation, yielding more stable and meaningful results even in lower-width regimes. However, pCNL is more computationally expensive and does not offer clear advantages in large-scale settings, making it less preferable in such cases.

\subsection{Future work}
While our study provides valuable insights into the effectiveness of the pCN and pCNL algorithms in sampling from wide BNNs, it is important to acknowledge several limitations that may affect the generalizability of our findings.

We limited our experiments to one-layer fully connected BNNs in order to isolate and highlight the impact of width changes on the performance of the samplers. More extensive experiments with deeper networks and different architectures are necessary to fully explore the applicability of these methods in real-world case studies. Testing the algorithms on a wider variety of datasets will help determine their robustness across diverse tasks.

In terms of future works, an interesting direction involves the analysis of our methods in conjunction with existing studies on the mixing speed of MCMC algorithms. Specifically, a key challenge in the field is to establish that the number of steps required for the marginal distribution of MCMC procedures to converge to the target distribution does not grow exponentially with the number of parameters. For the pCN algorithm, this question has been addressed positively in Theorem 2.12 of \cite{Hairer_2014}, where conditions under which the number of steps remains manageable are provided. Since, loosely speaking, the main assumption requires the acceptance rate to have a lower bound that is strictly positive, an application of the result in our framework seems possible. Indeed, in the context of our work, we have this assumption for granted by showing that the acceptance rate converges to 1 as the network width increases. Thus, it would be interesting to combine our results with these findings to explore further implications for chain mixing properties, potentially offering new insights into the efficiency of pCN and pCNL in high-dimensional settings. See also \cite{Andrieu_2024}.

\section*{Acknowledgments}

S.F. acknowledges the Italian Ministry of Education, University and Research (MIUR), “Dipartimenti di Eccellenza" grant 2023-2027. 
%\bibliography{references}

%%%%%%%%%%%%%%%%%%%%%%%%%%%%%%%%%%%%%%%%%%%%%%%%%%%%%%%%%%%%

\clearpage

\onecolumn
\appendix

\section{Proofs and derivations}

\subsection{Proof of Theorem 2.1}\label{proof_acc}
Let the assumptions of Theorem \ref{Thm1} hold. We start by analysing the general expression of the MCMC acceptance probability:
\begin{equation*}
a = \min\left\{1, \frac{p(\phi^* | \mathcal{D})q(\phi | \phi^*)}{p(\phi | \mathcal{D})q(\phi^* | \phi)}\right\}.
\end{equation*}
We have already shown that $q(\phi^* | \phi) = \mathcal{N}(\sqrt{1-\beta^2}\phi, \beta^2 I_D)$. Regarding the reparamatrized weight posterior of the network, we observe that \citet{hron_2022}
\begin{align*}
p(\phi | \mathcal{D}) & = p(\phi^{(L+1)} | \phi^{(\le L)}, \mathcal{D}) p(\phi^{(\le L)}| \mathcal{D})                                                               \\
                      & \propto  p(\phi^{(L+1)} | \phi^{(\le L)}, \mathcal{D}) \sqrt{det(\Sigma)} exp\left( \frac{1}{2}y^T(\sigma^2 I_n + \Psi \Psi^T)^{-1}y \right)
\end{align*}

where $p(\phi^{(L+1)} | \phi^{(\le L)}, \mathcal{D}) \sim \mathcal{N}(0, I_{d^{(L)}})$ is assured by the reparametrisation.
It is then crucial to recognize the empirical NNGP kernel $\hat{K}_{\sigma^2} = \sigma^2 I_n + \Psi \Psi^T$ \citep{RasmussenW06} and observe that $det(\Sigma) \propto det(\hat{K}_{\sigma^2})$.

Inserting everything in the expression of the acceptance probability we have:
\begin{align*}
    \frac{p(\phi^*| \mathcal{D})q(\phi | \phi^*)}{p(\phi| \mathcal{D})q(\phi^* | \phi)} &= \frac{p(\phi^{*(\le L)}| \mathcal{D})p(\phi^{*(L+1)}|\phi^{*(\le L)}, \mathcal{D})q(\phi | \phi^*)}{p(\phi^{(\le L)}| \mathcal{D})p(\phi^{(L+1)}|\phi^{(\le L)}, \mathcal{D})q(\phi^* | \phi)}\\
    &= \frac{p(\phi^{*(\le L)})\sqrt{det(\Sigma^*)} exp\left( \frac{1}{2}y^T(\sigma^2 I_n + \Psi^* \Psi^{*T})^{-1}y \right)p(\phi^{*(L+1)}|\phi^{*(\le L)}, \mathcal{D})q(\phi | \phi^*)}{p(\phi^{(\le L)})\sqrt{det(\Sigma)} exp\left( \frac{1}{2}y^T(\sigma^2 I_n + \Psi \Psi^T)^{-1}y \right)p(\phi^{(L+1)}|\phi^{(\le L)}, \mathcal{D})q(\phi^* | \phi)}
\end{align*}

Where we denote with $\Sigma^*$ and $\Psi^*$ the covariance matrix and scaled input matrix of the redout layer in equation \ref{cov_mean}, but for a network with weights $\phi^*$ .
Now:
\begin{align*}
    q(\phi|\phi^*) &\propto exp\left( -\frac{1}{2\beta^2} ||\phi- \sqrt{1-\beta^2}\phi^*||^2 \right)\\
    &= exp\left( -\frac{1}{2\beta^2} ||\phi||^2 -\frac{(1-\beta^2)}{2\beta^2}||\phi^*||^2 + \frac{\sqrt{1-\beta^2}}{\beta^2}\phi^T\phi^* \right)\\
    &=exp\left( -\frac{1}{2\beta^2} ||\phi||^2 -\frac{1}{2\beta^2} ||\phi^*||^2 + \frac{1}{2}||\phi^*||^2 + \frac{\sqrt{1-\beta^2}}{\beta^2}\phi^T\phi^* \right)
\end{align*}

From which
\begin{align*}
\frac{q(\phi|\phi^*)}{q(\phi^*|\phi)} = exp\left(\frac{1}{2} ||\phi^*||^2 - \frac{1}{2} ||\phi||^2 \right)
\end{align*}

and since
\begin{align*}
    p(\phi^{*(\le L)})p(\phi^{*(L+1)}|\phi^{*(\le L)}, \mathcal{D}) &\propto exp\left( -\frac{1}{2} ||\phi^{*(\le L)}||^2 \right)exp\left( -\frac{1}{2} ||\phi^{*(L+1)}||^2 \right)\\
    &= exp\left( -\frac{1}{2} ||\phi^*||^2 \right)
\end{align*}

we obtain
\begin{align*}
    \frac{p(\phi^*| \mathcal{D})q(\phi | \phi^*)}{p(\phi| \mathcal{D})q(\phi^* | \phi)} &= \frac{exp\left( -\frac{1}{2} ||\phi^*||^2 \right)exp\left(\frac{1}{2} ||\phi^*||^2 \right)\sqrt{det(\Sigma^*)} exp\left( \frac{1}{2}y^T(\sigma^2 I_n + \Psi^* \Psi^{*T})^{-1}y \right)}{exp\left( -\frac{1}{2} ||\phi||^2 \right)exp\left(\frac{1}{2} ||\phi||^2 \right)\sqrt{det(\Sigma)} exp\left( \frac{1}{2}y^T(\sigma^2 I_n + \Psi \Psi^T)^{-1}y \right)}\\
    &= \frac{\sqrt{det(\Sigma^*)} exp\left( \frac{1}{2}y^T(\sigma^2 I_n + \Psi^* \Psi^{*T})^{-1}y \right)}{\sqrt{det(\Sigma)} exp\left( \frac{1}{2}y^T(\sigma^2 I_n + \Psi \Psi^T)^{-1}y \right)}\\
    &\propto \frac{\sqrt{det(\hat{K}^*_{\sigma^2})} exp\left( \frac{1}{2}y^T(\hat{K}^*_{\sigma^2})^{-1}y \right)}{\sqrt{det(\hat{K}_{\sigma^2})} exp\left( \frac{1}{2}y^T(\hat{K}_{\sigma^2})^{-1}y \right)}
\end{align*}

To conclude, we exploit the known convergence of the empirical NNGP kernel to a constant independent of $\phi^{\le L} = \theta^{\le L}$
\begin{equation*}
\hat{K}_{\sigma^2} \to K_{\sigma^2} \quad \text{as } \, d_{min} \to \infty
\end{equation*}

This proves that the numerator and the denominator converge to the same quantity and, consequently, that their ratio converges to 1.

Implying the thesis
\begin{equation*}
a = 1 \wedge \frac{p(\phi^*| \mathcal{D})q(\phi | \phi^*)}{p(\phi| \mathcal{D})q(\phi^* | \phi)} \to 1 \wedge 1 = 1 \quad \text{for } d_{min} \to \infty
\end{equation*}

\subsection{Proof of Theorem 2.2}\label{proof_thm2}
The proof of Theorem \ref{Thm2} relies heavily on results obtained along the proof of Theorem \ref{Thm1}.

Let's start by writing expliciteply the expressions of the acceptance probability for the pCNL algorithm:
\begin{align*}
    \rho(\phi,\phi^*) &- \rho(\phi^*,\phi) =\\ 
    = - &\ell(\phi) - \frac{1}{2}\langle\phi^*-\phi, \mathcal{D}\ell(\phi)\rangle - \frac{\delta}{4} \langle\phi+\phi^*, \mathcal{D}\ell(\phi)\rangle + \frac{\delta}{4} \norm{\sqrt{C}\mathcal{D}\ell(\phi)}^2\\
     + &\ell(\phi^*) + \frac{1}{2}\langle\phi-\phi^*, \mathcal{D}\ell(\phi^*)\rangle + \frac{\delta}{4} \langle\phi+\phi^*, \mathcal{D}\ell(\phi^*)\rangle - \frac{\delta}{4} \norm{\sqrt{C}\mathcal{D}\ell(\phi^*)}^2\\
    = - &\ell(\phi) + \ell(\phi^*) - \frac{1}{2}\langle\phi-\phi^*, \mathcal{D}\ell(\phi) + \mathcal{D}\ell(\phi^*)\rangle\\ 
    - &\frac{\delta}{4} \langle\phi+\phi^*, \mathcal{D}\ell(\phi) - \mathcal{D}\ell(\phi^*)\rangle + \frac{\delta}{4} \left( \norm{\mathcal{D}\ell(\phi)}^2 - \norm{\mathcal{D}\ell(\phi^*)}^2\right)\\
\end{align*}

Now, we can exploit previous results to evaluate the likelihood distribution:

\begin{align*}
    exp&\{\ell(\phi)\} = \frac{p(\phi|\mathcal{D})p(\mathcal{D})}{p(\phi)} =\\
    =& \frac{C_1(\mathcal{D})p(\phi^{(L+1)}|\phi^{(\le L)}, \mathcal{D})p(\phi^{(\le L)})\sqrt{det(\Sigma)} exp\left( \frac{1}{2}y^T(\sigma^2 I_n + \varPsi \varPsi^T)^{-1}y \right)}{p(\phi)}\\
    =&C_1(\mathcal{D})\sqrt{det(\Sigma)} exp\left( \frac{1}{2}y^T(\sigma^2 I_n + \varPsi \varPsi^T)^{-1}y \right)
\end{align*}
where $C_1$ is a suitable distribution depending only on data. The last equality is due to the fact that 
$p(\phi^{(L+1)}|\phi^{(\le L)}, \mathcal{D})p(\phi^{(\le L)}) \propto exp\left( -\frac{1}{2}\norm{\phi}^2 \right)$ and the prior $p(\phi)$ is assumed to be a standard Gaussian.

\bigskip

As already done in the proof for pCN sampler we recognize the empirical NNGP kernel $\hat{K}_{\sigma^2}$, and we recall that as the minimum layer width goes to $\infty$, 
it converges to a constant independent of the weights $K_{\sigma^2}$.

Hence, noting that the Fréchet derivative is a bounded linear operator (thus continuous), by continuous mapping theorem we can conclude that $\mathcal{D}\ell(\phi) \to 0$ as the layer
width goes to $\infty$. Since the same goes for $\mathcal{D}\ell(\phi^*)$, we can conclude that in the wide-width limit, all the terms containing Fréchet derivatives of the 
log-likelihood, evaluated at any point, vanish. As a result, in the wide-width regime, the acceptance rate of the pCNL sampler converges to the one of pCN and, thus, converges to 1.

\subsection{Proof of Theorem 3.1}\label{proof_acc_margin}
Since we are sampling only from the inner-weights $\theta^{(\le L)} = W$ of the BNN, the acceptance probability becomes

\begin{align*}
a = 1 \wedge \frac{p(W | \mathcal{D})q(W | W^*)}{p(W | \mathcal{D})q(W^* | W)}
\end{align*}

By definition \ref{repar}, $p(W | \mathcal{D}) = p(\phi^{(\le L)} | \mathcal{D})$ and thus we can exploit the known expressions from Section \ref{proof_acc}:
\begin{align*}
    p(W| \mathcal{D}) &=  p(W)\sqrt{det(\Sigma)} exp\left( \frac{1}{2}y^T(\sigma^2 I_n + \Psi \Psi^T)^{-1}y \right)\\
    &\propto exp\left(-\frac{1}{2} ||W||^2 \right)\sqrt{det(\Sigma)} exp\left( \frac{1}{2}y^T(\sigma^2 I_n + \Psi \Psi^T)^{-1}y \right)
\end{align*} 
Moreover
\begin{align*}
q(W & |W^*) \propto exp\left( -\frac{1}{2\beta^2}|| W - \sqrt{1 - \beta^2}W^* ||^2 \right) \\ &\implies \frac{q(W | W^*)}{q(W^* | W)} = exp\left( \frac{1}{2}||W^*||^2 - \frac{1}{2}||W||^2 \right)
\end{align*}

Putting all together and simplifying we get:

\begin{align*}
    a &= 1 \wedge \frac{\pi(W^* | \mathcal{D})q(W | W^*)}{\pi(W | \mathcal{D})q(W^* | W)}\\
    &= 1 \wedge \frac{\sqrt{det(\Sigma^*)} exp\left( \frac{1}{2}y^T(\sigma^2 I_n + \Psi^* \Psi^{*T})^{-1}y \right)}{\sqrt{det(\Sigma)} exp\left( \frac{1}{2}y^T(\sigma^2 I_n + \Psi \Psi^T)^{-1}y \right)}
\end{align*}
That corresponds to the same exact expression found in Section
\ref{proof_acc} and hence convergence to 1 as the layers width increases is granted by the same arguments.

\section{Additional experiments}
\label{additional_experiments}
\subsection{Gelman-Rubin Statistic Analysis}
\label{gelman_rubin}
The Gelman-Rubin statistic or potential scale reduction factor, $\hat{R}$, \citep{gelman1992inference, brooks1998general} complements the ESS analsys by assessing how the different samples affect the rate of convergence of the collected Markov chain. More precisely, the metric is evaluated by running multiple independent chains per each sample and then exploiting comparisons of the variance between multiple chains to the variances within the single chains to give a measure of the degree to which variance (of the means) between chains exceeds what one would expect if the chains were identically distributed.
Mathematically, the Gelman-Rubin statistic is defined as follows:
Suppose we have run $M$ chains in parallel with different starting values and that we have collected $N$ samples for each chain. Denote by $x_i^{(m)}$ the $i$-th sample of the 
$m$-th chain. The within-chain variance and the between-chain variance are defined as:
\begin{equation*}
    W = \frac{1}{M} \sum_{m=1}^{M} s_m^2, \quad\quad B = \frac{N}{M-1} \sum_{m=1}^{M} (\bar{x}^{(\cdot)} - \bar{x}^{(m)})^2,
\end{equation*}
where $s_m^2$ is the sample variance of the $m$-th chain, $\bar{x}^{(m)}$ its mean and $\bar{x}^{(\cdot)}$ is the grand mean.
The Gelman-Rubin statistic is then defined as:
\begin{equation*}
    \hat{R} = \frac{\frac{N-1}{N}W+\frac{1}{N}B}{W}.
\end{equation*}

Now, since at convergence all the parallel chains are indistinguishable, the Gelman-Rubin statistic should be close to 1. In practice, a value of $\hat{R}$ smaller than $1.2$ is considered as a good indicator of approximate convergence \citep{gelman1992inference}.

In all the following experiments we make use of the TensorFlow Probability’s built-in function, \texttt{tfp.mcmc.potential\_scale\_reduction}, to compute $\hat{R}$.
\begin{figure*}[htbp]
    \centering
    \includegraphics[width=0.9\textwidth]{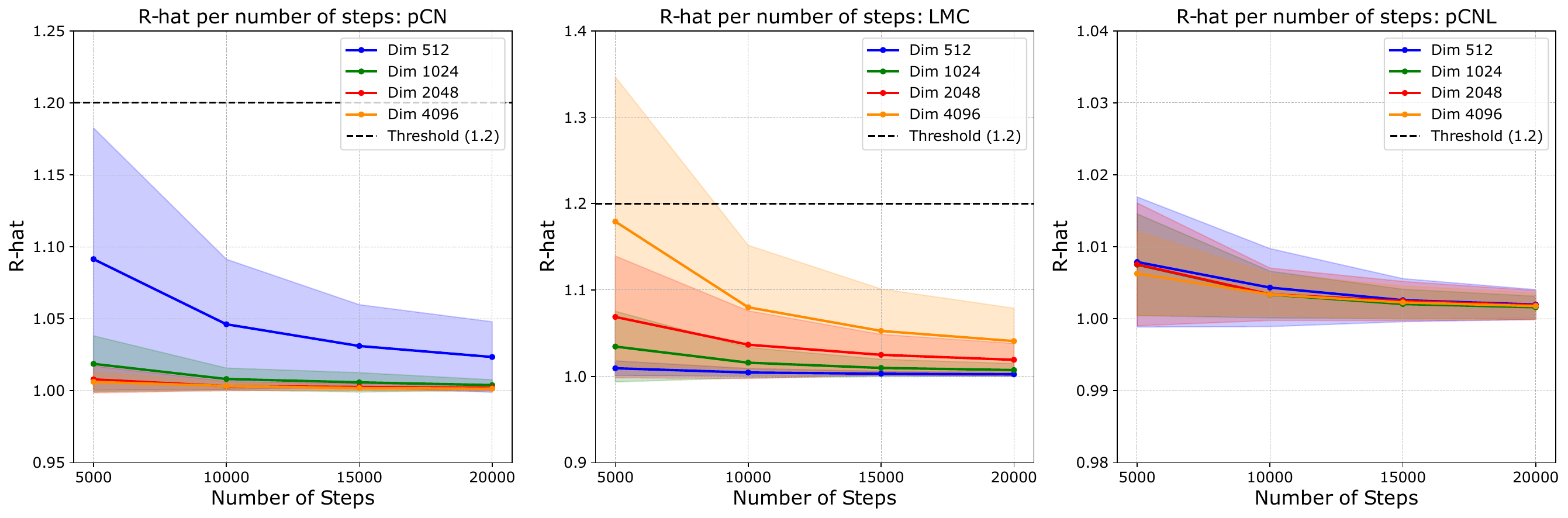}
    \caption{Evolution of the Gelman-Rubin statistic, evaluated using three independent chains, of the LMC, pCN and pCNL as a function of the number of steps for the 
    stepsize $\beta = 0.1$. The solid lines represent the average Gelman-Rubin statistic, whereas the shaded areas indicate their standard deviation. Again, the 
    {\fontfamily{qcr}\selectfont CIFAR-10} dataset, with sample size fixed at $n=256$, is used. Since the metric should be close to 1 for all chains to be considered converged,
    we trace a horizontal line at $\hat{R} = 1.2$, indicating the standard empirical threshold for determining convergence. 
    For all samples the chosen burn-in of 20,000 steps appears to be sufficient to ensure that the chains have reached the target distribution. The metric for pCN sampler improves as the width increases, whereas the LMC method shows complementary results. Finally, the pCNL approach exhibits consistent and 
    good results among all dimensions.}
    \label{fig:gelman_rubin}
\end{figure*}

We examine the $\hat{R}$ values for the pCN, LMC, and pCNL samplers, considering their evolution as a function of the number of steps 
[5000, 10000, 15000, 20000]. The stepsize is fixed at $\beta = 0.1$.
For each sampler, we run three independent chains for each of the BNN widths among $\{512, 1024, 2096, 4192\}$, we then use the collected samples to calculate the metric 
values. More specifically, our focus is on investigating how the $\hat{R}$ values for different widths of the BNN change as the number of steps increases for the different 
samplers. This analysis allows us to evaluate and compare the convergence efficiency of the different samplers across these varying network sizes.  The results are presented 
in Figure \ref{fig:gelman_rubin}, where we also plot the reference line of $\hat{R} = 1.2$ to indicate the standard empirical threshold for determining convergence.

In the first place we observe that for all samplers the chosen burn-in of 20,000 steps appears to be adequate to ensure that the chains have reached stationarity. Nevertheless,
the samplers present some peculiarities that are worth to discuss. The pCN sampler, shows a less variable and closer to 1 Gelman-Rubin statistic as the network width increases,
suggesting that the chain attains stationarity quicker and more steadily for large network widths. This is in line with the previously obtained results and 
confirms once again pCN suitability in wide BNNs settings. Not surprisingly, the LMC results seems the negative version of the pCN ones. Indeed, its $\hat{R}$ values reflects
the better performance of the sampler in lower dimensions, as underlined by its increase in both the average and standard deviation as the width grows. Finally, the pCNL
procedure exhibits exceptionaly good results among all dimensions, with the $\hat{R}$ values that are always close to 1 and have small variability. This is a very 
interesting result, since it highlights the ability of the sample to properly explore the space and quickly approach the target distribution.

\subsection{Trace plots of iterates PCs}
\label{trace_plots}
To further assess the convergence behavior of the MCMC samplers, we analyze the trace plots of the principal components (PCs) of the collected iterates. These plots visualize the evolution of the dominant modes of variation in the sampled posterior over time, providing insight into whether the Markov chain has reached stationarity. Specifically, well-mixed chains should exhibit stable oscillations around a mean value, without strong trends or prolonged correlations. Below, we present the trace plots of the first two principal components for all the considered sampling methods and stepsizes.

\begin{figure}[ht]
    \centering
    \includegraphics[trim={0 0 0 0cm},clip,width=0.9\textwidth]{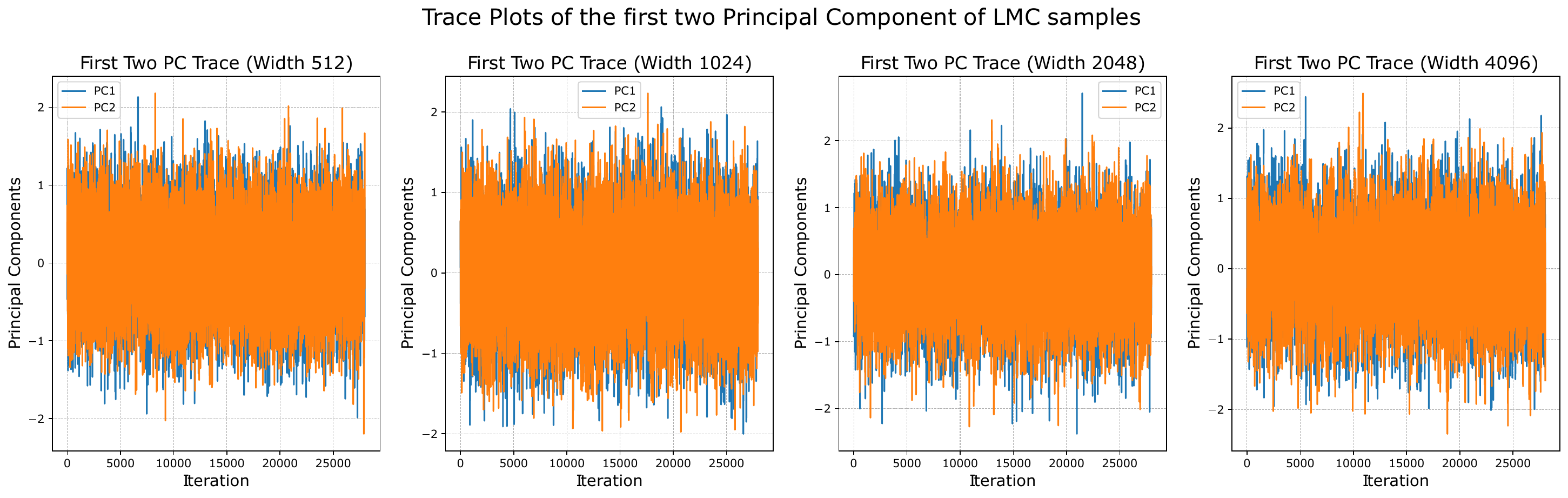}
    \includegraphics[trim={0 0 0 2.25cm},clip,width=0.9\textwidth]{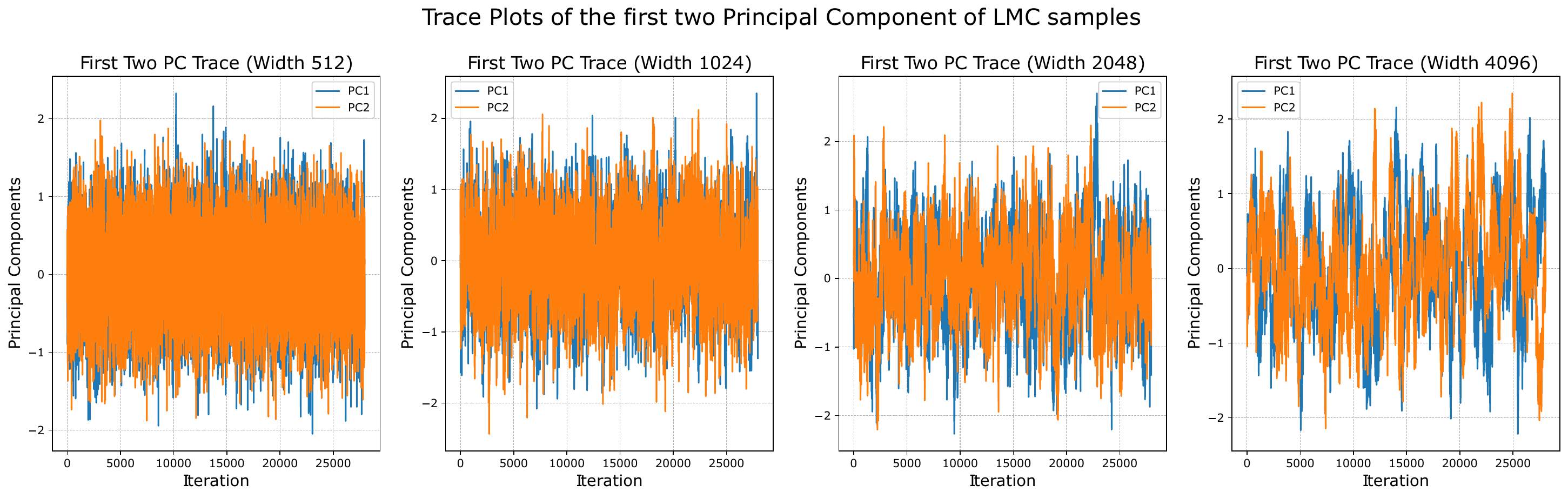}
    \includegraphics[trim={0 0 0 2.25cm},clip,width=0.9\textwidth]{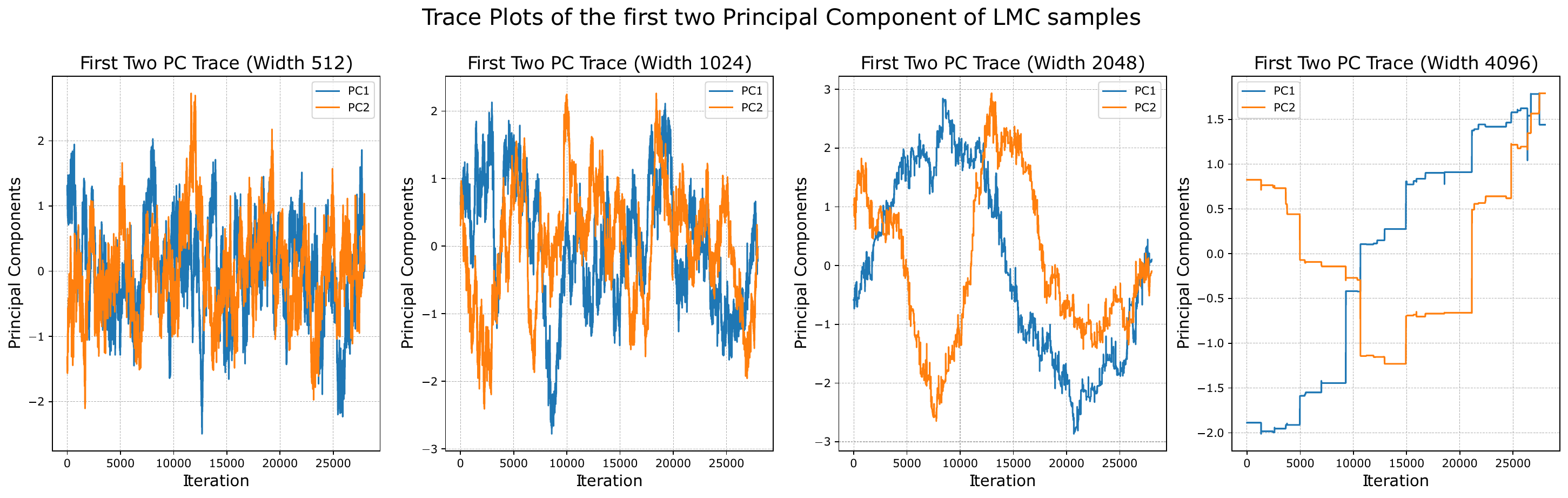}
    \caption{Trace plots of the first two principal components for the LMC sampler with different stepsizes. From top to bottom: $\beta = 0.01$, $\beta = 0.1$, and $\beta = 0.2$. A well-mixed chain should exhibit stable oscillations around a mean value, without strong trends or correlations.}
    \label{fig:trace_lmc}
\end{figure}

We can notice in Figure \ref{fig:trace_lmc} how the LMC sampler mixing properties deteriorate with the layers width, reflecting its well-known sensitivity to high-dimensional settings.

\begin{figure}[ht]
    \centering
    \includegraphics[trim={0 0 0 0cm},clip,width=0.9\textwidth]{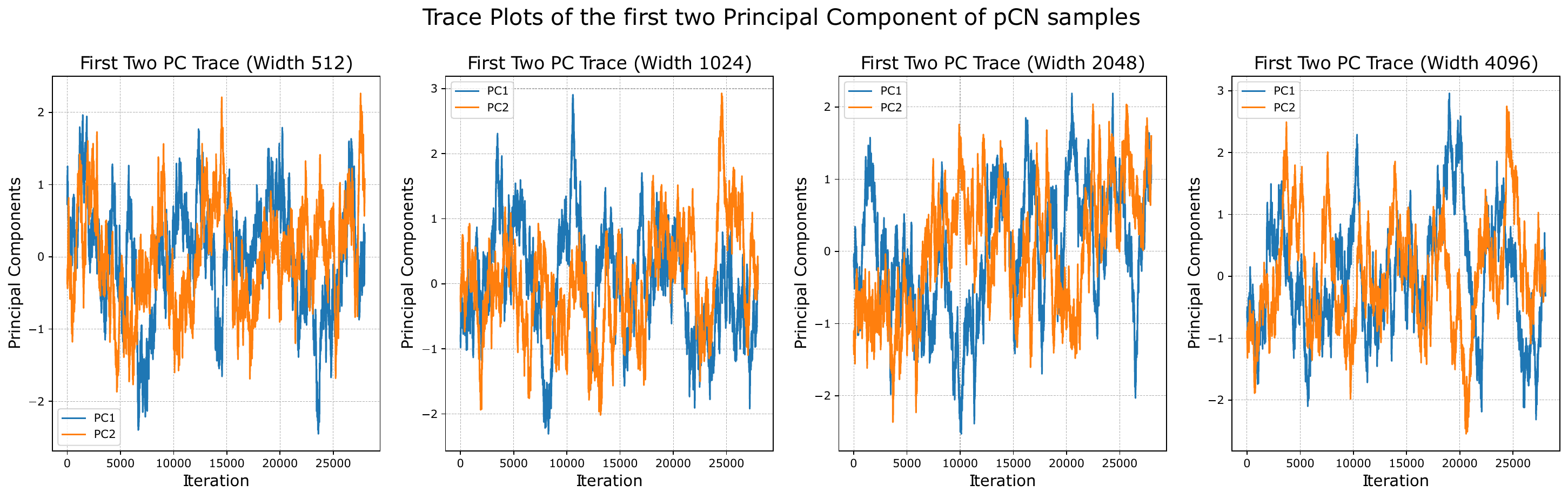}
    \includegraphics[trim={0 0 0 2.25cm},clip,width=0.9\textwidth]{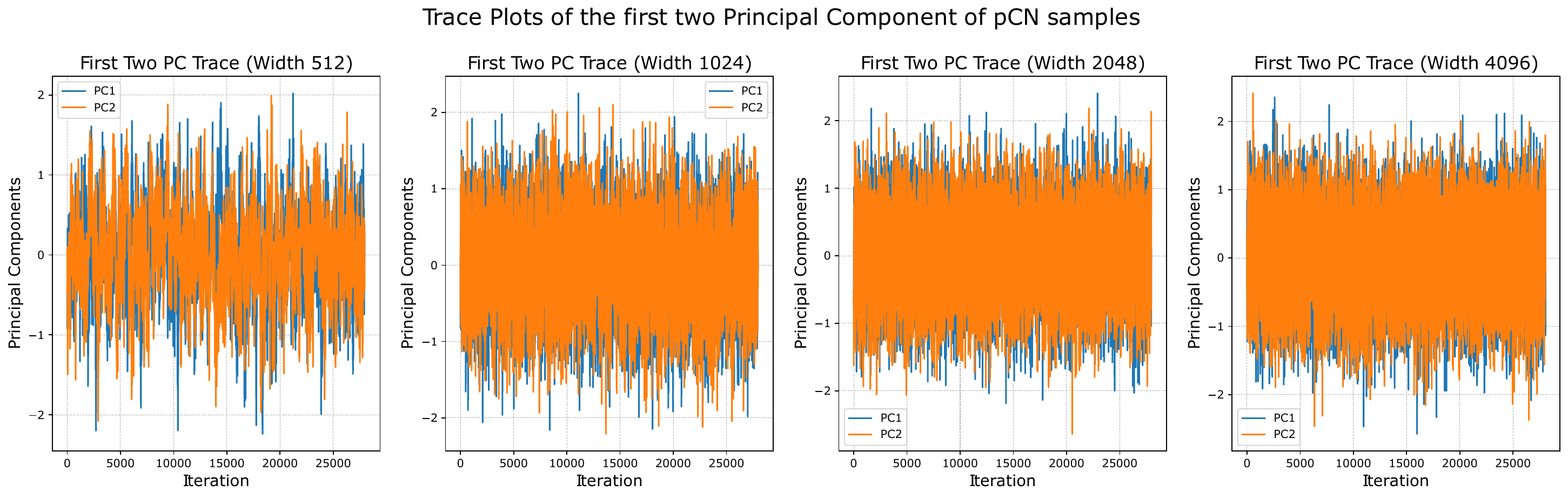}
    \includegraphics[trim={0 0 0 2.25cm},clip,width=0.9\textwidth]{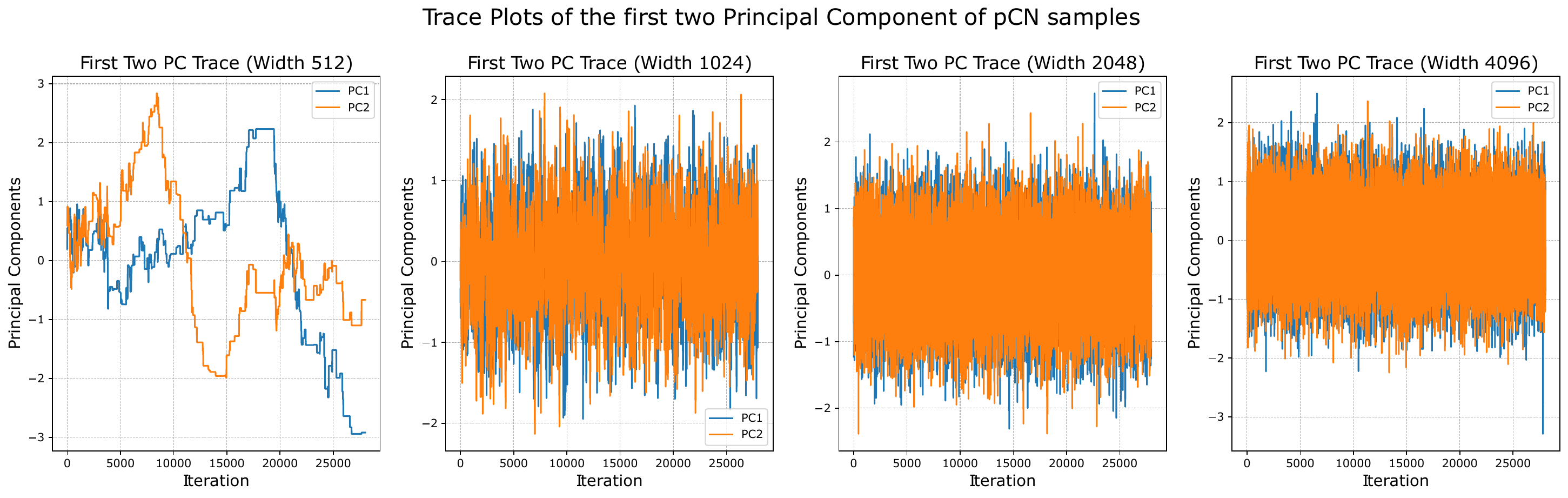}
    \caption{Trace plots of the first two principal components for the pCN sampler with different stepsizes. From top to bottom: $\beta = 0.01$, $\beta = 0.1$, and $\beta = 0.2$. The stability of the trace plots improves for larger network widths, demonstrating the suitability of pCN in wide BNNs.}
    \label{fig:trace_pcn}
\end{figure}

The pCN sampler, Figure \ref{fig:trace_pcn}, shows the opposite behavior: as the network width increases, the trace plots become more stable, supporting the theoretical guarantee that its acceptance rate converges to 1 in the infinite-width regime.

\begin{figure}[ht]
    \centering
    \includegraphics[trim={0 0 0 0cm},clip,width=0.9\textwidth]{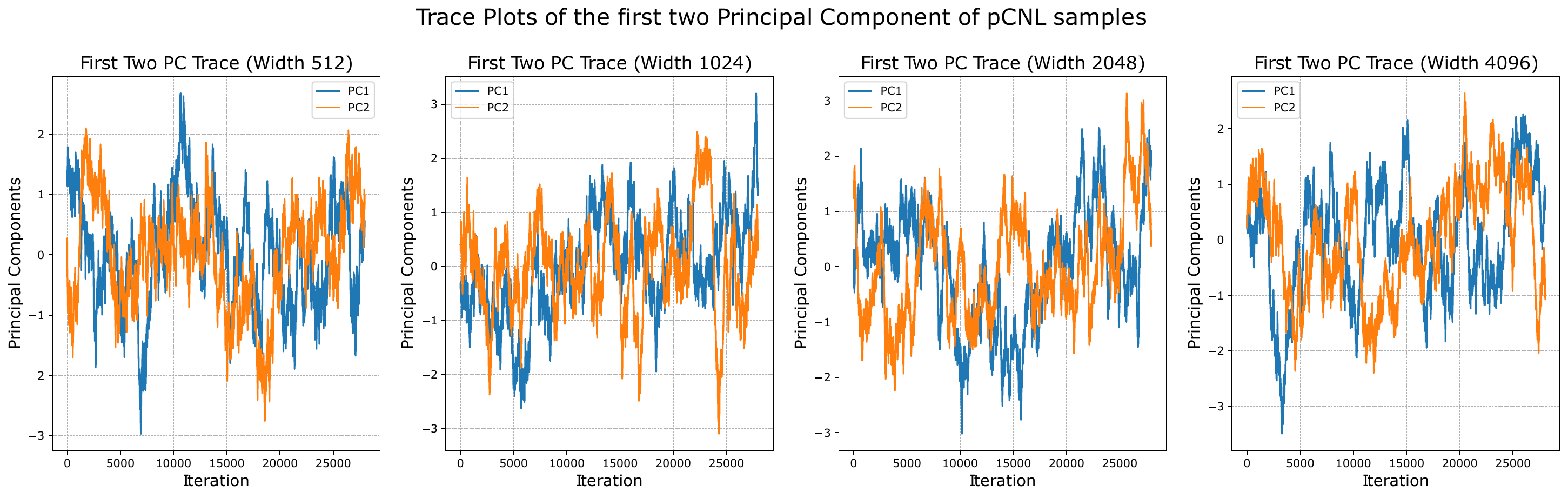}
    \includegraphics[trim={0 0 0 2.25cm},clip,width=0.9\textwidth]{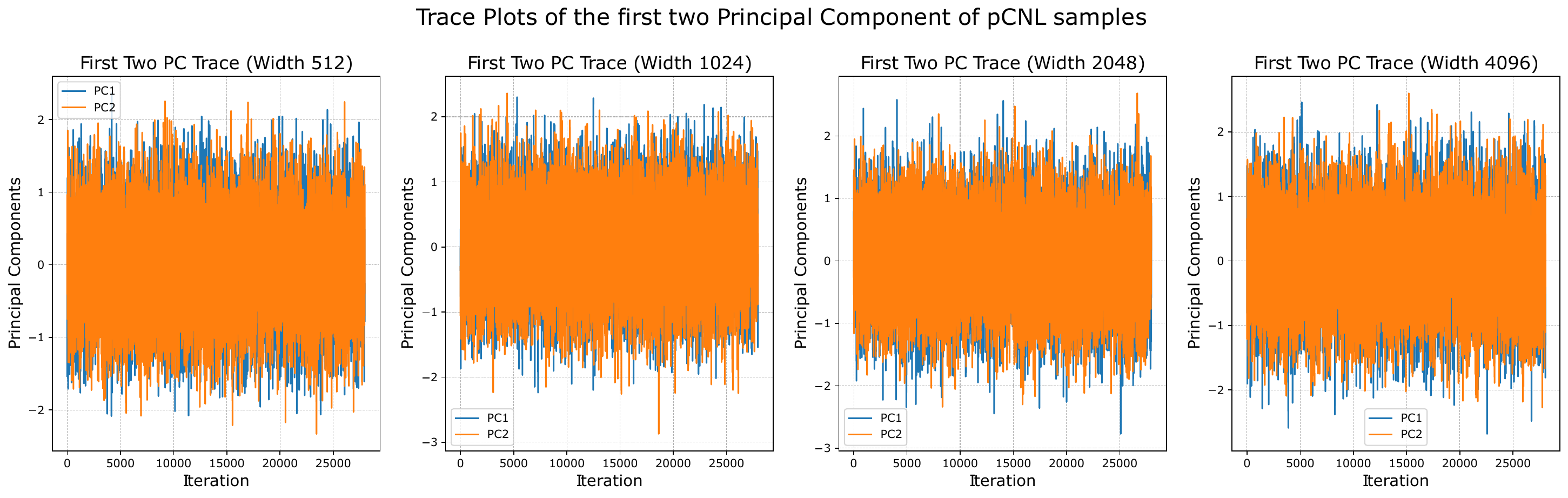}
    \includegraphics[trim={0 0 0 2.25cm},clip,width=0.9\textwidth]{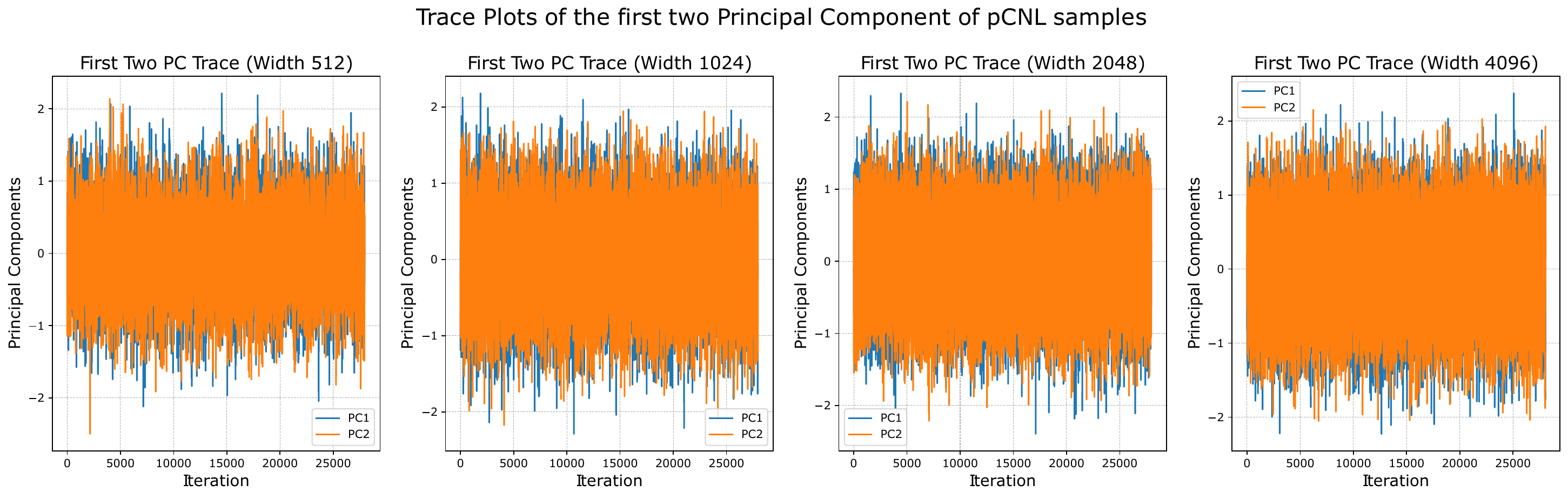}
    \caption{Trace plots of the first two principal components for the pCNL sampler with different stepsizes. From top to bottom: $\beta = 0.01$, $\beta = 0.1$, and $\beta = 0.2$. The performance of pCNL is analyzed to understand its convergence properties in comparison to LMC and pCN.}
    \label{fig:trace_pcnl}
\end{figure}

Finally, in Figure \ref{fig:trace_pcnl}, appears that the pCNL samples maintains good mixing properties also in the smaller width regimes.

\section{Further mathematical foundations and justifications}

\subsection{Fréchet derivative}
\label{frechet}

The Fréchet derivative is a generalization of the derivative to function spaces. Given a Banach space \( X \) and a function \( F: X \to Y \) mapping between Banach spaces, we say that F is Fréchet differentiable at a point \( \theta \in X \) if there is a bounded linear operator \( DF(\theta): X \to Y \) satisfying:

\begin{equation}
    \lim_{\|h\| \to 0} \frac{\| F(\theta + h) - F(\theta) - DF(\theta)[h] \|_Y}{\|h\|_X} = 0.
\end{equation}

We call \( DF(\theta) \) the Fréchet derivative of F at $\theta$.

This definition ensures that \( DF(\theta) \) provides the best linear approximation of \( F \) around \( \theta \), analogous to the Jacobian in finite-dimensional spaces.

The Fréchet derivative of the log-likelihood plays a crucial role in the definition of the pCNL sampler, as it directly influences the proposal dynamics. Here we add details on its computation. Given the likelihood function:

\begin{equation}
    p(y | \theta, x) \sim \mathcal{N}(f^{(L+1)}(x), \sigma^2 I),
\end{equation}

the corresponding log-likelihood function is:

\begin{equation}
    \ell(\theta) = -\frac{n}{2} \log(2\pi\sigma^2) - \frac{1}{2\sigma^2} \sum_{i=1}^{n} \| y_i - f^{(L+1)}(x_i) \|^2.
\end{equation}

To compute the Fréchet derivative, we differentiate \(\ell(\theta)\) with respect to \(\theta\). Since \( f^{(L+1)}(x) \) depends on the network parameters \( \theta \), we apply the chain rule:

\begin{equation}
    D\ell(\theta) = -\frac{1}{\sigma^2} \sum_{i=1}^{n} (y_i - f^{(L+1)}(x_i)) \nabla_{\theta} f^{(L+1)}(x_i),
\end{equation}

where \( \nabla_{\theta} f^{(L+1)}(x) \) is the Jacobian of the network output with respect to the parameters \( \theta \), computed recursively using backpropagation.

\end{document}